\documentclass[runningheads]{llncs}
\usepackage[T1]{fontenc}
\usepackage{graphicx}
\usepackage{subcaption}
\usepackage{xspace}
\usepackage{amsmath}

\usepackage{dsbda-style}

\usepackage{booktabs}
\usepackage{threeparttable}
\usepackage[nolist]{acronym}

\usepackage{caption}
\usepackage{tabularx}
\usepackage{multirow}
\usepackage{booktabs}
\usepackage{setspace}
\usepackage{makecell}
\usepackage{dblfloatfix}
\usepackage[utf8]{inputenc}
\usepackage[font=small,labelfont=bf]{caption}

\begin{document}
\begin{acronym}
\acro{BERT}[BERT]{Bidirectional Encoder Representations from Transformers}
\acro{MLM}[MLM]{Masked Language Modeling}
\acro{NSP}[NSP]{Next Sentence Prediction}
\acro{RoBERTa}[RoBERTa]{Robustly optimized BERT approach}
\acro{BPE}[BPE]{Byte-Pair Encoding}
\acro{DeBERTa}[DeBERTa]{De\-coding-enhanced BERT with disentangled attention}
\acro{ReLU}[ReLU]{Rectified Linear Unit}
\acro{MLP}[MLP]{Multilayer Perceptron}
\acro{SOTA}[SOTA]{State of the Art}
\acro{LSTM}[LSTM]{Long short-term memory}
\acro{DADGNN}[DADGNN]{Deep Attention Diffusion Graph Neural Net\-work}
\acro{GNN}[GNN]{Graph Neural Network}
\acro{ALBERT}[ALBERT]{A Lite BERT}
\acro{SOP}[SOP]{Sentence-Order Prediction}
\acro{WIPO}[WIPO]{World Intellectual Property Organization}
\acro{STOPS}[STOPS]{Short Texts Of Products and Services}
\acro{CNN}[CNN]{Convolutional Neural Network}
\acro{RNN}[RNN]{Recurrent Neural Networks}
\acro{SGNN}[SGNN]{Sequential Graph Neural Net\-work}
\acro{BoW}[BoW]{Bag of Words}
\acro{InducT-GCN}[InducT-GCN]{InducTive Graph Convolutional Networks for Text classification}
\acro{GCN}[GCN]{Graph Convolutional Networks}
\acro{NLP}[NLP]{Natural Language Processing}
\acro{HGAT}[HGAT]{Heterogeneous Graph Attention networks}
\acro{HIN}[HIN]{Heterogeneous Information Network}
\acro{LDA}[LDA]{Latent Dirichlet Allocation}
\acro{POS}[POS]{Part Of Speech}
\acro{STGCN}[STGCN]{Short-Text \acl{GCN}}
\end{acronym}

\title{Transformers are Short-text Classifiers}
\titlerunning{Transformers are Short-text Classifiers}
\author{Fabian Karl
\orcidID{0009-0008-0079-5604} 
\and
Ansgar Scherp
\orcidID{0000-0002-2653-9245}
}
\institute{Universität Ulm, Germany
\email{\{fabian.karl, ansgar.scherp\}@uni-ulm.de}}
\maketitle              

\begin{abstract}
Short text classification is a crucial and challenging aspect of Natural Language Processing. For this reason, there are numerous highly specialized short text classifiers. A variety of approaches have been employed in short text classifiers such as convolutional and recurrent networks. Also many short text classifier based on graph neural networks have emerged in the last years. However, in recent short text research, State of the Art (SOTA) methods for traditional text classification, particularly the pure use of Transformers, have been unexploited. In this work, we examine the performance of a variety of short text classifiers as well as the top performing traditional text classifier on benchmark datasets. We further investigate the effects on two new real-world short text datasets in an effort to address the issue of becoming overly dependent on benchmark datasets with a limited number of characteristics. The datasets are motivated from a real-world use case on classifying goods and services for tax auditing. NICE is a classification system for goods and services that divides them into 45 classes and is based on the Nice Classification of the World Intellectual Property Organization. The Short Texts Of Products and Services (STOPS) dataset is based on Amazon product descriptions and Yelp business entries. Our experiments unambiguously demonstrate that Transformers achieve SOTA accuracy on short text classification tasks, raising the question of whether specialized short text techniques are necessary. The NICE dataset showed to be particularly challenging and makes a good benchmark for future advancements.

Source code is available here: \url{https://github.com/FKarl/short-text-classification}

\keywords{Text Classification \and Transformer \and BERT \and GNN}
\end{abstract}

\section{Introduction}
\label{sec:introduction}

Text classification is a crucial aspect of \ac{NLP}, and extensive research in this field is being conducted. Many researchers are working to improve the speed, accuracy, or robustness of their algorithms.
Traditional text classification, however, does not take some traits into account that appear in numerous real-world applications, such as short text.
Therefore, studies have been conducted specifically on short texts~\cite{shine,zhao2021sequential}. 
From user-generated content like social media to business data like accounting records, short text covers a wide range of topics.
For example, the division into goods and services (see Section~\ref{ŝec:new-datasets}) is an important part of the tax audit. 
Currently, an auditor checks whether the element descriptions match the appropriate class of good or service. 
Since this can be very time-consuming, it is desirable to bring it into a semi-automatic context with the help of classifiers. 
Also, the subdivision into more specific classes can be useful for determining whether a given amount for an entry in the accounting records is reasonable.

Since short texts are typically only one to two sentences long, they lack context and therefore pose a challenge for text classification.
In order to get better results, many short text classifiers also operate in a transductive setup~\cite{hgat,shine,ye2020document}, which includes the test set during training. 
However, as they need to be retrained each time new data needs to be classified, those transductive models are not very suitable for real-world applications.
The results of both transductive and the generally more useful inductive short text classifier are typically unsatisfactory due to the challenge that short text presents.
Recent studies on short texts have emphasized specialized models~\cite{hgat,nc-hgat,zhao2021sequential,wang2021short,shine,ye2020document} to address the issues associated with the short text length.
However, \acf{SOTA} text classification methods, particularly the pure use of Transformers, have been unexploited.
In this work, the effectiveness on short texts is examined and tested by means of benchmark datasets.
We also introduce two new, realistic datasets in the domain of goods and services descriptions.
Our contributions are in summary:
\begin{itemize}
    \item We provide a comparison of various modern text classification techniques.
    In particular, specialized short text methods are compared with the top performing traditional text classification models.
    
    \item We introduce two new real-world datasets in the goods and services domain to cover additional dataset characteristics in a realistic use-case.

    \item Transformers achieve SOTA accuracy on short text classification tasks. 
    This questions the need of specialized short text classifier.
\end{itemize}

Below, we summarize the related work.
Section~\ref{sec:models} provides a description of the models that were selected for our experiments.
The experimental apparatus is described in Section~\ref{sec:experiments}.
An overview of the achieved results is reported in Section~\ref{sec:results}. 
Section~\ref{sec:discussion} discusses the results, before we conclude.

\section{Related Work}
\label{sec:related_work}
Despite the fact that \ac{BoW}-based models have long represented the cutting edge in text classification, attention has recently shifted to sequence-based and, more recently, graph-based concepts.
However, \ac{BoW}-based models continue to offer a solid baseline~\cite{galkeMLP}. 
For example in fastText~\cite{joulin-etal-2017-fasttext} the average of the trained word representations are used as text representation and then fed into a linear classifier. This results in an efficient model for text classification.
To give an overview of the various concepts, Section~\ref{rw-sequence} provides various works in the field of sequence-based models, Section~\ref{rw-graph} discusses graph-based models, and Section~\ref{rw-short} examines  how these concepts are applied to short text. Finally, a summary of the findings from the related work is presented in Section~\ref{rw-summary}.

\subsection{Sequence-based Models}
\label{rw-sequence}

For any NLP task, \ac{RNN} and \ac{LSTM} are frequently used and a logical choice because both models learn historical information while taking location information for all words into account \cite{survey,DBLP:journals/corr/abs-1801-06717}.
Since \acp{RNN} must be computed sequentially and cannot be computed in parallel, the use of \acp{CNN} is also common \cite{survey,oreillyTransformer}. The text must be represented as a set of vectors that are concatenated into a matrix in order to be used by \acp{CNN}. 
The standard CNN convolution and pooling operations can then be applied to this matrix.
TextCNN~\cite{kim_TextCNN} uses this in combination with pretrained word embeddings for sentence-level classification tasks.
While CNN-based models extract the characteristics from the convolution kernels, the relationship between the input words is captured by RNN-based models \cite{survey}.
An important turning point in the advancement of \ac{NLP} technologies was the introduction of \ac{BERT}~\cite{vaswani2017attention}.
By performing extensive pre-training in an unsupervised manner and automatically mining semantic knowledge, \ac{BERT} learns to produce contextualized word vectors that have a global semantic representation.
\extended{BERT-like models are suited for large datasets as they can parallelize computation \cite{survey}.}
The effectiveness of BERT-like models for text classification is demonstrated by Galke and Scherp~\cite{galkeMLP}.
\extended{However, very few works take them into account when classifying texts.}

\subsection{Graph-based Models}
\label{rw-graph}
Recently, text classification has paid a lot of attention to graph-based models, particularly \acp{GNN}~\cite{deng2022aggnn,wang2022induct,shi2022ilgcn}.
This is due to the fact that tasks with rich relational structures benefit from the powerful representation capabilities of GNNs, which preserve global structure information \cite{wang2022induct}.
The task of text classification offers this rich relational structure because text can be modeled as edges and nodes in a graph structure.
There are different ways to represent the documents in a graph structure, but two main approaches have emerged \cite{wang2022induct,shine}.
The first approach builds a graph for each document using words as nodes and structural data, such as word co-occurence data, as edges. However, only local structural data is used.
The task is constructed as a whole graph classification problem in order to classify the text.
A popular \emph{document-level} approach is HyperGAT~\cite{ding-etal-2020-hypergat} which uses a dual attention mechanism and hypergraphs applied to documents to learn text embeddings.
The second approach creates a graph for the entire corpus using words and documents as nodes.
The text classification task is now a node classification task for the unlabeled document nodes.
The drawback of this method is that models using it are inherently transductive. 
For example, TextGCN~\cite{TextGCN_Yao_Mao_Luo_2019} uses this concept by employing a standard \ac{GCN} on this heterogeneous graph.
Following TextGCN, Lin et al.~\cite{lin2021bertgcn} propose BertGCN, a model that makes use of BERT to initialize representations for the document nodes in order to combine the benefits of both the large-scale pretraining of BERT and the transductive TextGCN.
However, the increase provided by this method is limited to datasets with long average text lengths.
Zeng et al.~\cite{zeng2022simplified} also experiment with combining TextGCN and BERT in the form of TextGCN-Bert-serial-SB, a Simplified-Boosting Ensemble, where BERT is only trained on the TextGCN's misclassification.
Which model is applied to which document is determined by a heuristic based on the node degree of the test document.
However, TextGCN-CNN-serial-SB, which substitutes TextCNN for BERT, yields better results.
By using a joint training mechanism, TextING~\cite{zhang2020TextING} and BERT are trained on sub-word tokens and base their predictions on the results of the two models. 
In contrast to applying each model separately, this produces better results.
Another approach combining graph classifiers with BERT is ContTextING~\cite{huang-etal-2022-contexting}. 
ContTextING utilizes a joint training mechanism to create a unified model that incorporates both document-wise contextual information from a BERT-style model and node interactions within a document through the use of a GNN module. The predictions for the text classification task are determined by combining the output from both of these modules.

\subsection{Short Text Models}
\label{rw-short}
Of course, short texts can also be classified using the methods discussed above. 
However, this is challenging because short texts tend to lack context and adhere to less strict syntactic structure \cite{shine}.
This has led to the emergence of specialized techniques that focus on improving the results for short text.
Early works focused on sentence classification using methods like Support Vector Machines (SVM)~\cite{DBLP:journals/air/SilvaCMW11}.
A survey by Galke et al.~\cite{DBLP:conf/kcap/GalkeMSBS17} compared SVM and other classical methods like Naive Bayes and $k$NN with multi-layer perceptron models (MLP) on short text classification.
Other works on sentence classification used Convolutional Neural Networks (CNN)~\cite{wang2021short,DBLP:conf/naacl/ZhangRW16,DBLP:conf/acl/KalchbrennerGB14}, which showed strong performance on benchmark datasets.
Recently, also methods exploiting graph neural networks were adopted to the needs of short text.
For instance, \ac{HGAT}~\cite{hgat} is a powerful semi-supervised short text classifier.
This was the first attempt to model short texts as well as additional information like topics gathered from a \ac{LDA}~\cite{2003LDA} and entities retrieved from Wikipedia with a \ac{HIN}.
To achieve this, a HIN embedding with a dual-level attention mechanism for nodes and their relations was used. For the semantic sparsity of short text, both the additional information and the captured relations are beneficial.
A transductive and an inductive HGAT model were released, with the transductive model being better on every dataset.
\extended{By adapting neighbor contrastive learning from Hu et al.~\cite{graph-mlp},~}NC-HGAT~\cite{nc-hgat} expands the HGAT model to produce a more robust variant.
Neighbor contrastive learning is based on the premise that documents that are connected have a higher likelihood of sharing a class label and, as a result, should therefore be closer in feature space.
In order to represent the additional information, SHINE~\cite{shine} also makes use of a heterogenous graph. In contrast, SHINE generates component graphs in the form of word, entity, and \ac{POS} graphs and creates a dynamically learned short document graph by employing hierarchical pooling over all component graphs. In the semi-supervised setting, SHINE outperforms HGAT as a strong transductive model.
SimpleSTC (Simple Short Text Classification)~\cite{DBLP:conf/emnlp/ZhengWYD22} is a graph-based method for short-text classification similar to SHINE. 
But instead of constructing the word-graph only over the data corpus itself, SimpleSTC employs a global corpus to create a reference graph that shall enrich and help to understand the short text in the smaller corpus.
As global corpus, articles from Wikipedia are used.
The authors sample 20 labeled documents per class as training set and validation set.
\ac{STGCN}~\cite{ye2020document} is an additional short text classifier. A graph of topics, documents, and unique words is the foundation of STGCN. Although the STGCN results by themselves are not particularly strong, the impact of pre-trained word vectors obtained by BERT was also examined. The classification of the STGCN is significantly enhanced by the combination of STGCN with BERT and a Bi-LSTM.
\subsection{Summary}
\label{rw-summary}
Graph neural network-based methods are widely used in short text classification. 
However, in recent short text research, \ac{SOTA} text classification methods, particularly the pure use of Transformers, have been unexploited.
\extended{Even though STGCN uses BERT to enhance their outcomes, the results of their fine-tuned BERT suggests that they did not fully utilize BERT's potential.} 
The majority of short text models are transductive. 
The crucial drawback of being transductive is that every time new data needs to be classified, the model must be retrained.

\section{Selected Models for Our \extended{Text Classification }Comparison}
\label{sec:models}

\extended{We provide a description of the conceptual designs of the inductive models that were selected for our experiments.
We focus on inductive models since it resembles a more realistic real-world scenario.}
We begin with models for short text classification in Section~\ref{sec:models_short} and then Section~\ref{sec:models_long} introduces a selection of top-performing models for text classification.
Following Galke and Scherp~\cite{galkeMLP}, we have excluded works that employ non-standard datasets only, use different measures, or are otherwise not comparable.
For example, regarding short text classification there are works that are applied on non-standard datasets only~\cite{HuEtAl2022,DBLP:conf/nlpcc/ZhongZZZ21}.

\subsection{Models for Short Text Classification}
\label{sec:models_short}
The models listed below either make claims about their ability to categorize short texts or were designed with that specific goal.\extended{

\subsubsection*{SECNN}}
The \textbf{SECNN}~\cite{wang2021short} is a text classification model built on \acp{CNN} that was created specifically for short texts with few and insufficient semantic features.
Wang et al.~\cite{wang2021short} suggested four components to address this issue. In order to achieve better coverage on the word vector table, they used an improved Jaro-Winkler similarity during preprocessing to identify any potential spelling mistakes. Second, they use a \ac{CNN} model built on the attention mechanism to look for words that are related. 
Third, in order to accomplish the goal of short text semantic expansion, the external knowledgebase \emph{Probase}~\cite{probase} is used to enhance the semantic features of short text. Finally, the classification process is performed using a straightforward \ac{CNN} with  a Softmax output layer.

\extended{\subsubsection*{SGNN/ESGNN/C-BERT}}
The Sequential Graph Neural Network (\textbf{SGNN})~\cite{zhao2021sequential} is a \ac{GNN}-based model that emphasizes the propagation of features based on sequences. 
By training each document as a separate graph, it is possible to learn the words' local and sequential features. GloVe's~\cite{pennington2014glove} pre-trained word embedding is utilized as a semantic feature of words. In order to update the feature matrix for each document graph, a Bi-LSTM is used to extract the contextual feature of each word. After that, a simplified \ac{GCN} aggregates the features of neighboring word nodes.
Additionally, Zhao et al.~\cite{zhao2021sequential} introduce two variants: 
Extended-SGNN (\textbf{ESGNN}), in which the initial contextual feature of words is preserved, and \textbf{C-BERT}, in which the Bi-LSTM is swapped for BERT.

\extended{\subsubsection*{DADGNN}}
The Deep Attention Diffusion Graph Neural Network (\textbf{DADGNN})~\cite{liu2021deep} is a graph-based method that combats the oversmoothing problem of \acp{GNN} and allows stacking more layers by utilizing attention diffusion and decoupling techniques.
This decoupling technique is also very advantageous for short texts because it obtains distinct hidden features in deep graph networks.

\extended{\subsubsection*{Bi-LSTM}}
The \acf{LSTM}~\cite{hochreiter1997long}, which is frequently used in text classification, has a bidirectional variant called \textbf{Bi-LSTM}~\cite{DBLP:journals/corr/LiuQH16}.
Due to its strong results for short texts~\cite{zhao2021sequential,DBLP:journals/corr/abs-1801-06717} and years of use as the \ac{SOTA} method for many tasks, this model is a good baseline for our purpose. 

\subsection{Top-performing Models for Text Classification}
\label{sec:models_long}
An overview of the top text classification models that excel on texts of all lengths and were not specifically created with short texts in mind is provided in this section.
We employ the base models for the Transformers.

\extended{\subsubsection*{BERT}}
\label{sec:BERT}
The Bidirectional Encoder Representations from Transformers (\textbf{BERT})~\cite{devlin2018bert} is a language representation model that is based on the Transformer architecture~\cite{vaswani2017attention}. Encoder-only models, such as \ac{BERT}, rely solely on the encoder component of the Transformer architecture, whereby the text sequences are converted into rich numerical representations~\cite{oreillyTransformer}. These models are well suited for text classification due to this representation.
\ac{BERT} is designed to incorporate a token's left and right contexts into its computed representation. This is commonly referred to as bidirectional attention.
\extended{For pre-training, \ac{BERT} uses two unsupervised objectives: 

(1)~\ac{MLM}:
    In this method, a percentage of the input tokens are masked at random with a [MASK] token, and the training objective is to predict those masked tokens. 
    In order to counteract the effect of causing a mismatch between pre-training and fine-tuning, ``masked tokens'' are not always replaced by [MASK].
    The ``Masked tokens'' are $80\%$ of the time replaced by [MASK],  $10\%$ of the time replaced by a random token, $10\%$ of the time left unmodified.
    
(2)~\ac{NSP}:
    The process of determining whether a given sentence is the next sentence after another is known as next sentence prediction.
    The input for this objective is a pair of two sentences separated by a [SEP] token, with $50\%$ of the training data consisting of random second sentences and not the subsequent one. This aims to improve understanding of the relationship between two sentences, which language modeling does not directly capture.}

\extended{\subsubsection*{RoBERTa}}
The Robustly optimized BERT approach (\textbf{RoBERTa})~\cite{liu2019roberta} is a systematically improved \ac{BERT} adaptation.
In the \ac{RoBERTa} model, the pre-training strategy was changed and training was done on larger batches with more data, to increase \ac{BERT}'s performance. 

\extended{The following are the differences from \ac{BERT}:
(1)~Dynamic masking:
    While masking was done once during preprocessing in \ac{BERT}, \ac{RoBERTa} masks sequences dynamically as they are fed into the model. This is significant because static masking requires duplicated training data to obtain different maskings, which is impractical for large data sets.
(2)~FULL-SENTENCES without NSP loss:
    Instead of using the NSP task, RoBERTa fills the input with full sentences that are sampled coherently until the total length reaches $512$ tokens to learn long-range dependencies.
(3)~Large mini-batches:
    Liu et al.~\cite{liu2019roberta} notice that training with large batches enhances perplexity and end-task accuracy for the masked language modeling target. Therefore, they raise the batch size to 8K. 
(4)~Larger byte-level \ac{BPE}:
    BERT's 30K BPE vocabulary is replaced by a larger byte-level BPE vocabulary with 50K subword units, following the idea of Radford et al.~\cite{radford2019language}. This adds approximately 15M parameters and decrease end-task performance slightly, but it provides the benefit of a universal encoding scheme.
    }

\extended{\subsubsection*{DeBERTa}}
To improve BERT and RoBERTa models, Decoding-enhanced BERT with disentangled attention (\textbf{DeBERTa})~\cite{DBLP:conf/iclr/HeLGC21} makes two architectural adjustments.
The first is the disentangled attention mechanism, which encodes the content and location of each word using two vectors. The content of the token at position $i$ is represented by  $H_i$ and the relative position $i | j$ between the token at position $i$ and $j$ are represented by $P_{i|j}$. The equation for determining the cross attention score is as follows:
 $    A_{i,j} = H_iH_j^T + H_iP_{j|i}^T + P_{i|j}H_j^T + P_{i|j}P_{j|i}^T$.
The second adjustment is an enhanced mask decoder that uses absolute positions in the decoding layer to predict masked tokens during pre-training. 
For masked token prediction, DeBERTa includes the absolute position after the transform layers but before the softmax layer.
In contrast, BERT incorporates the position embedding into the input layer.
As a result, DeBERTa is able to capture the relative position in all Transformer layers.

\extended{\subsubsection*{ERNIE 2.0}}
Sun et al.~\cite{ERNIE} proposed \textbf{ERNIE 2.0}, a continuous pre-training framework that builds and learns pre-training tasks through continuous multi-task learning. 
This allows the extraction of additional valuable lexical, syntactic, and semantic information in addition to co-occurring information, which is typically the focus.

\extended{\subsubsection*{DistilBERT}}
The concept behind \textbf{DistilBERT}~\cite{DISTILBERT} is to leverage knowledge distillation to produce a more compact and faster version of BERT while retaining most of its language understanding capacities. DistilBERT reduces the size of BERT by $40\%$, is $60\%$ faster, and still retains $97\%$ of its language understanding capabilities.
In order to accomplish this, DistilBERT optimizes the following three objectives while using the BERT model as a teacher:
(1)~Distillation loss:
    The model was trained to output probabilities equivalent to those of the BERT base model. 
    \extended{
    The loss function is defined as cross-entropy over the soft target probabilities
    \begin{equation*}
        L_{ce} = \sum_i t_i \cdot \log(s_i)
    \end{equation*}
    where $t_i$ and $s_i$ are the probabilities estimated by the teacher and the student, respectively.}
(2)~\acf{MLM}:
    As described by Devlin et al.~\cite{devlin2018bert} for the BERT model, the common pre-training using masked language modeling is being used\extended{ (see Section~\ref{sec:BERT})}.
(3)~Cosine embedding loss: The model was trained to align the DistilBERT and BERT hidden state vectors\extended{ as close as possible}.

\extended{\subsubsection*{ALBERTv2}}
A Lite BERT (\textbf{ALBERT})~\cite{albert2019} is a Transformer that uses two parameter-reduction strategies to save memory and speed up training by sharing the weights of all layers across its Transformer. 
This model is therefore particularly effective for longer texts. 
During pretraining, ALBERTv2 employs \ac{MLM} and \ac{SOP}, which predicts the sequence of two subsequent text segments.

\extended{\subsubsection*{WideMLP}}
\textbf{WideMLP}~\cite{galkeMLP} is a \ac{BoW}-Based \ac{MLP} with a single wide hidden layer of $1,024$ \acp{ReLU}. 
This model serves as a useful benchmark against which we can measure actual scientific progress.

\extended{\subsubsection*{InducT-GCN}}
InducTive Graph Convolutional Networks for Text classification (\textbf{InducT-GCN})~\cite{wang2022induct} is a \ac{GCN}-based method that categorically rejects any information or statistics from the test set. 
To achieve the inductive setup, InducT-GCN represents document vectors with a weighted sum of word vectors and applies TF-IDF weights instead of representing document nodes with one-hot vectors.
A two-layer GCN is employed for training, with the first layer learning the word embeddings and the second layer in the dimension of the dataset's classes outputs into a softmax activation function. 

\section{Experimental Apparatus}
\label{sec:experiments}
\extended{First, the datasets used in our study are presented. 
Then we describe the preprocessing steps and our procedure. 
The selection and optimization of our hyperparameters are covered next. Finally, the experimentation's measures will be described.}

\subsection{Datasets}
\extended{We employed a variety of datasets, which are further explained in this section, to demonstrate the performance of the various short text classification models.}
First, we describe the benchmark datasets. Second, we introduce our new datasets in the domain of goods and services. 
The characteristics are denoted in Table~\ref{tab:datasets}.
    \begin{table}[ht]
       \small
        \centering
        \caption{Characteristics of short text datasets. 
        \extended{We use benchmark datasets as well as introduce new datasets for Goods \& Services.}
        \#C refers to the number of classes.
        Avg. L is the average document length.
        }\label{tab:datasets}
        \begin{tabular}{l|rrrrr}

\toprule
            \textbf{Benchmarks}          & \textbf{\#Doc} & \textbf{\#Train} & \textbf{\#Test} & \textbf{\#C} & \textbf{Avg. L} \\
            \midrule
            R8                                   & 7,674          & 5,485            & 2,189           & 8                  & 65.72                \\
            MR                                   & 10,662         & 7,108            & 3,554           & 2                  & 20.39                \\
            SearchSnippets                       & 12,340         & 10,060           & 2,280           & 8                  & 18.10                \\
            Twitter                              & 10,000         & 7,000            & 3,000           & 2                  & 11.64                \\
            TREC                                 & 5,952          & 5,452            & 500             & 6                  & 10.06                \\
            SST-2                                & 9,613          & 7,792            & 1,821           & 2                  & 20.32                \\

            \toprule
            \textbf{Goods \& Services}           & \textbf{\#Doc} & \textbf{\#Train} & \textbf{\#Test} & \textbf{\#C} & \textbf{Avg. L} \\
            \midrule
            NICE-45                              & 9,593          & 6,715            & 2,878           & 45                 & 3.75                 \\
            NICE-2                               & 9,593          & 6,715            & 2,878           & 2                  & 3.75                 \\
            STOPS-41                             & 200,341        & 140,238          & 60,103          & 41                 & 5.64                 \\
            STOPS-2                              & 200,341        & 140,238          & 60,103          & 2                  & 5.64                 \\
            \bottomrule
        \end{tabular}
    \end{table}

\subsubsection{Benchmark Datasets}
Six short text benchmark datasets, namely R8, MR, SearchSnippets, Twitter, TREC, and SST-2, are used in our experiments. The following gives a detailed description of them.
\textbf{R8} is an 8-class subset of the Reuters 21578 news dataset\footnote{\url{http://www.daviddlewis.com/resources/testcollections/reuters21578/}}. It is not a classical short text scenario with an average length of $65.72$ tokens but offers the ability to set the methods in comparison to traditional text classification.
\textbf{MR}\footnote{\url{https://www.cs.cornell.edu/people/pabo/movie-review-data/}} is a widely used dataset for text classification. It contains movie-review documents with an average length of $20.39$ tokens and is therefore suitable for short text classification.
The dataset \textbf{SearchSnippets}\footnote{\url{http://jwebpro.sourceforge.net/data-web-snippets.tar.gz}}, which is made up of snippets returned by a search engine and has an average length of $18.10$ tokens, was released by Phan et al.~\cite{phan2008learning}.
\textbf{Twitter}\footnote{\url{https://www.nltk.org/howto/twitter.html\#Using-a-Tweet-Corpus}} is a collection of $10,000$ tweets that are split into the categories negative and positive based on sentiment. The length of those tweets is on average $11.64$ tokens.
\textbf{TREC}\footnote{\url{https://cogcomp.seas.upenn.edu/Data/QA/QC/}}, which was introduced by Li and Roth~\cite{li-roth-2002-learning}, is a question type classification dataset with six classifications for questions. It provides the shortest texts in our collection of benchmark datasets, with an average text length of $10.06$ tokens.
\textbf{SST-2}\footnote{\url{https://nlp.stanford.edu/sentiment/}}~\cite{SocherEtAl2013:RNTN} or SST-binary is a subset of the Stanford Sentiment Treebank, a fine-grained sentiment analysis dataset, in which neutral reviews have been removed and the data has either a positive or negative label. 
    The average number of tokens in the texts is $20.32$.

\begin{figure}[htbp]
\centering

\begin{subfigure}[t]{0.24\textwidth}
  \includegraphics[width=\textwidth]{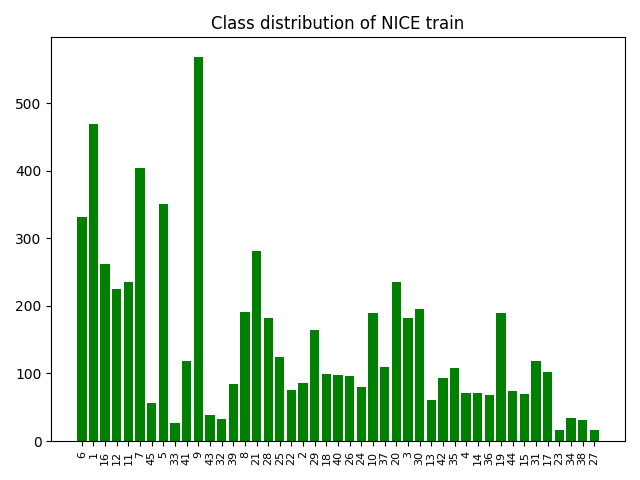}
  \caption{NICE - train}
\end{subfigure}
\hfill
\begin{subfigure}[t]{0.24\textwidth}
  \includegraphics[width=\textwidth]{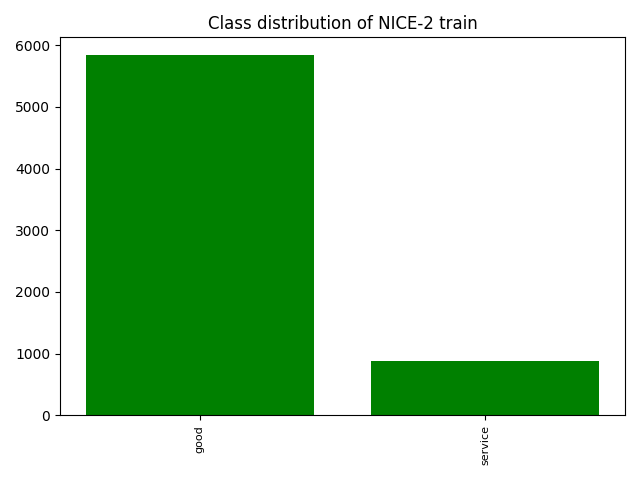}
  \caption{NICE-2 - train}
\end{subfigure}
\hfill
\begin{subfigure}[t]{0.24\textwidth}
  \includegraphics[width=\textwidth]{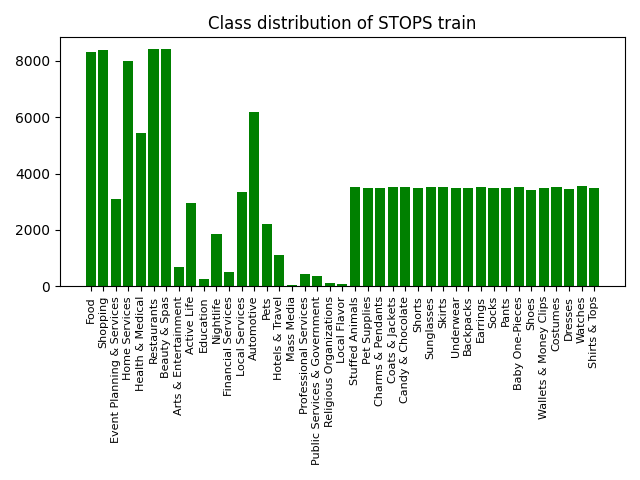}
  \caption{STOPS - train}
\end{subfigure}
\hfill
\begin{subfigure}[t]{0.24\textwidth}
  \includegraphics[width=\textwidth]{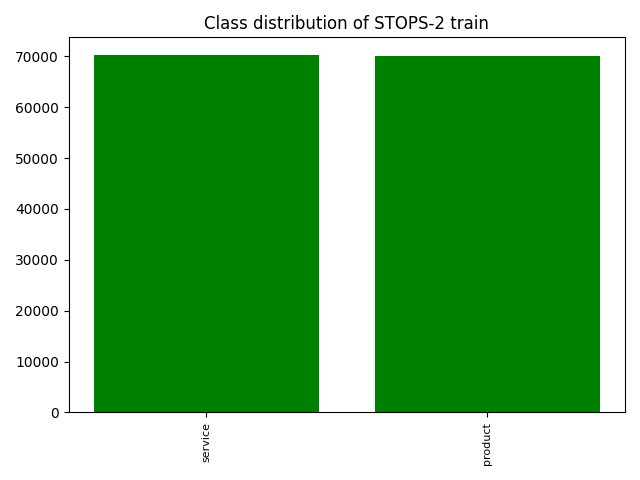}
  \caption{STOPS-2 - train}
\end{subfigure}

\begin{subfigure}[t]{0.24\textwidth}
  \includegraphics[width=\textwidth]{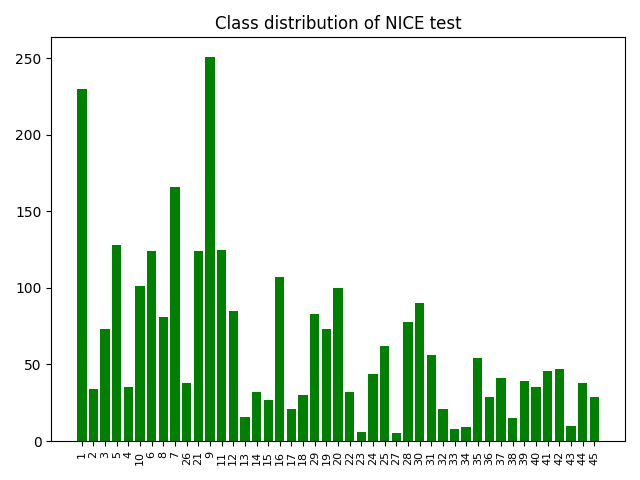}
  \caption{NICE - test}
\end{subfigure}
\hfill
\begin{subfigure}[t]{0.24\textwidth}
  \includegraphics[width=\textwidth]{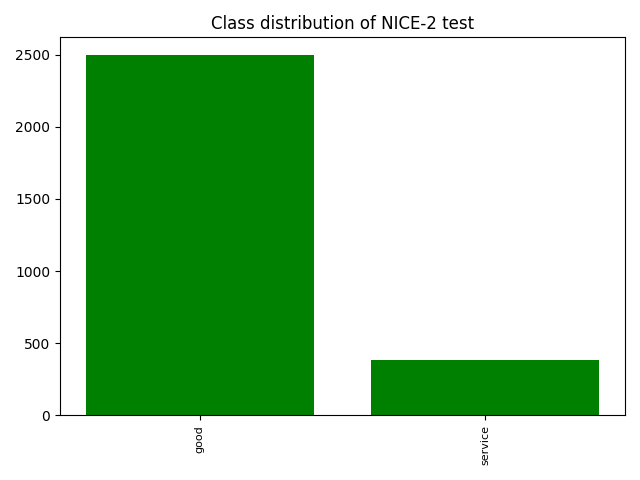}
  \caption{NICE-2 - test}
\end{subfigure}
\hfill
\begin{subfigure}[t]{0.24\textwidth}
  \includegraphics[width=\textwidth]{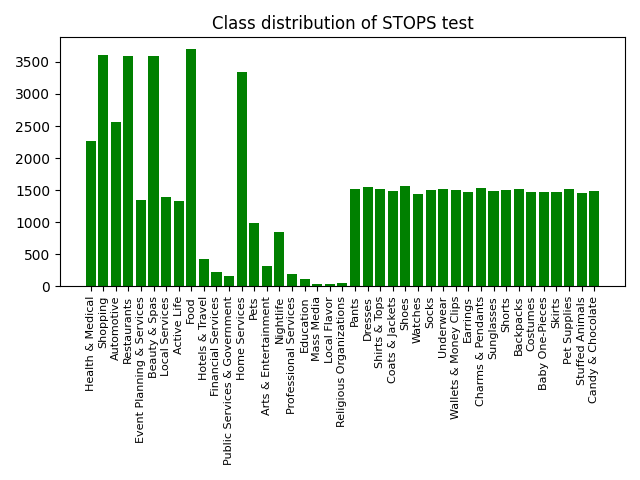}
  \caption{STOPS - test}
\end{subfigure}
\hfill
\begin{subfigure}[t]{0.24\textwidth}
  \includegraphics[width=\textwidth]{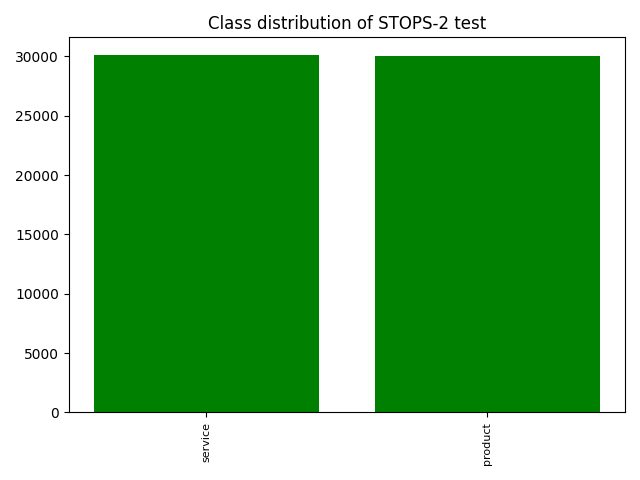}
  \caption{STOPS-2 - test}
\end{subfigure}

\caption{Class distribution of our new datasets (separated by train and test split)}
\label{fig:dataset_statistics}
\end{figure}

\subsubsection{Goods and Services Datasets}
\label{ŝec:new-datasets}
In order to evaluate the performance on data with real world applications, we introduce two new datasets that are focused on the distinction between goods and services.
Although there are already datasets for product classification, such as the WDC-LSPM\footnote{\url{http://webdatacommons.org/largescaleproductcorpus/}}, to the best of our knowledge, our datasets are the first to combine goods and services.
\textbf{NICE} is a classification system for goods and services that divides them into $45$ classes and is based on the Nice Classification\footnote{\url{https://www.wipo.int/classifications/nice/en/}} of the \ac{WIPO}. There are $11$ classes for various service types and $34$ categories for goods.
With $9,593$ documents, NICE-45 is comparable in size to the benchmark datasets.
This dataset, which has texts with an average length of $3.75$ tokens, is an excellent example of extremely short text. 
For the division into goods and services, there is also the binary version NICE-2.
\textbf{\ac{STOPS}} is the second dataset we offer. With $200,341$ documents and an average length of $5.64$ tokens, STOPS-41 is a reasonably large dataset. The data set was derived from a potential use case in the form of Amazon descriptions and Yelp business entries, making it the most realistic. Like NICE, STOPS has a binary version STOPS-2.
Both datasets provide novel characteristic properties that the benchmark datasets did not cover. In particular, the number of fine-granular classes presents a challenge that is not addressed by common benchmarks. For details on the class distribution of these datasets, please refer to Figure \ref{fig:dataset_statistics}.

\todo{MR also has a fine grained variant, TREC as far as I know too}

\subsection{Preprocessing}
\label{sec:preprocessing}
\extended{The preprocessing used to produce NICE and STOPS will be covered in this section.}
To create NICE, the \ac{WIPO}\footnote{\url{https://www.wipo.int/nice/its4nice/ITSupport\_and\_download\_area/20220101/MasterFiles/index.html}} classification data was converted to lower case, all punctuation was removed, and side information that was enclosed in brackets was also removed. Additionally, accents were dropped.
Following a random shuffle, the data was divided into $70\%$ train and $30\%$ test.

As product and service entries for STOPS, we use the product descriptions of MAVE\footnote{\url{https://github.com/google-research-datasets/MAVE}}~\cite{yang2022mave} and the business names of YELP\footnote{\url{https://www.yelp.com/dataset/download}}.
Due to the different data sources, these also had to be preprocessed differently.
All classes' occurrences in the MAVE data were counted, and $5,000$ sentences from each of the $20$ most common classes were chosen.
The multi-label categories for the YELP data were broken down into a list of single label categories, and the sentences were then mapped to the most common single label that each one has. In order to prevent any label from taking up too much of the dataset, the data was collected such that there is a maximum of $1,200$ documents per label.
After that, all punctuation was dropped, the data was converted to lower case, and accents were also dropped. The data was split into train and test in a $70:30$ ratio after being randomly shuffled.

\subsection{Procedure}
The best short text classifier and text classification models were retrieved from the literature (see description of the models in Section~\ref{sec:models}).
The accuracy scores were extracted in order to establish a comparison.
Own experiments, particularly using various Transformers, were conducted in order to compare them. Investigations into the impacts of hyperparameters on short texts were performed. More details about these are provided in Section~\ref{sec:hyperparameter}.
In order to test the methods in novel contexts, we also created two new datasets, whereby \ac{STOPS} stands out due to its much higher quantity of documents.

\subsection{Hyperparameter Optimization}
\label{sec:hyperparameter}

Our experiments for BERT, DistilBERT, and WideMLP used the hyperparameter from Galke and Scherp~\cite{galkeMLP}.
The parameters for BERT and DistilBERT are a learning rate of $5 \cdot 10^{-5}$, a batch size of $128$, and fine-tuning for $10$  epochs.
WideMLP was trained for $100$ epochs with a learning rate of $10^{-3}$, a batch size of $16$, and a dropout of $0.5$.
For ERNIE 2.0 and ALBERTv2, we make use of the SST-2 values that Sun et al.~\cite{ERNIE} and Lan et al.~\cite{albert2019}, respectively, published.
For our hyperparameter selection for DeBERTa and RoBERTa, we used the BERT values from Galke and Scherp~\cite{galkeMLP} as a starting point and investigated the effect of smaller learning rates.
\extended{The hyperparameters are chosen using the Bayesian hyperparameter search method, which models the relationship between the parameters and accuracy using a Gaussian process.
We chose intervals of the best-performing parameters retrieved from the Bayesian search method in order to obtain uniform parameters for all datasets. Our hyperparameters are then defined by the median of the intersection of these intervals. In our case, none of the intersections were empty.}
This resulted in learning rates of $2 \cdot 10^{-5}$ for DeBERTa and $4 \cdot 10^{-5}$ for RoBERTa while maintaining the other parameters.
For comparison, we followed the same procedure to create ERNIE 2.0~(optimized), which yields a learning rate of $25 \cdot 10^{-6}$.
The Bi-LSTM values from Zhao et al.~\cite{zhao2021sequential} were used for both the LSTM and the Bi-LSTM model.
We used DADGNN with the default parameters of $0.5$ dropout, $10^{-6}$ weight decay, and two attention heads for all datasets.

\subsection{Metrics}
\label{sec:metrics}
Accuracy is used to measure the classification of short text. 
\extended{The formula is as follows: 

\begin{equation}
  \label{eq:accuracy}
  Accuracy = \frac{TP + TN}{TP + TN + FP + FN}  
\end{equation}

where TP is the number of positive classes and classified to be positive classes, TN is the number of negative classes and classified to be negative classes.
FP and FN are the numbers of wrongly classified positive or negative classes, respectively.}
For multi-class cases, the subset accuracy is calculated.
\todo{@Ansgar: Source for the subset accuracy: \url{https://scikit-learn.org/stable/modules/generated/sklearn.metrics.accuracy_score.html}}

\section{Results}
\label{sec:results}

\todo{fix table and mention sd (5 runs) for induct-gcn and shine in caption?. Maybe a seperate table for the sd?

}
\todo{decide what to do with SD}
\todo{@Ansgar; Is it intended only SearchSni in the table? Snippets would be a more common abbreviation}

\renewcommand{\textsubscript}[1]{}
\begin{table*}[!ht]
    \caption[Accuracy on short text classification datasets]{
    \centering
    Accuracy on short text classification datasets. The ``Short?'' column indicates whether the model makes claims about its ability to categorize short texts. Provenance refers to the source of the accuracy scores.
    }\label{tab:accuracy}
    \centering

    \begin{threeparttable} 
        \scriptsize
        \resizebox{1.26\linewidth}{!}{

        \begin{tabular}{lcllllllr}

            \toprule
            \textbf{Inductive Models}               & \textbf{Short?} & \textbf{R8}      & \textbf{MR}    & \textbf{Snippets}           & \textbf{Twitter} & \textbf{TREC} & \textbf{SST-2} & \textbf{Provenance}\\
            \midrule
            \emph{Transformer Models} \\
            BERT                                    & N                   & 98.17\tnote{$1$} & 86.94          & 88.20                             & 99.96            & \textbf{99.4} & 91.37          & Own experiment                  \\
            RoBERTa                                 & N                   & 98.17\tnote{$1$} & 89.42          & 85.22                             & 99.9             & 98.6          & 94.01          & Own experiment                  \\
            DeBERTa                                 & N                   & 98.45\tnote{$1$} & \textbf{90.21} & 86.14                             & 99.93            & 98.8          & \textbf{94.78} & Own experiment                  \\
            ERNIE 2.0                               & N                   & 98.04\tnote{$1$} & 88.97          & 89.12                             & \textbf{99.97}   & 98.8          & 93.36          & Own experiment                  \\
            ERNIE 2.0 (optimized)                   & N                   & 98.17\tnote{$1$} & 89.53          & 89.17                             & \textbf{99.97}   & 99            & 94.07          & Own experiment                  \\
            DistilBERT                              & N                   & 97.98\tnote{$1$} & 85.31          & 89.69                             & 99.96            & 99            & 90.49          & Own experiment                  \\
            ALBERTv2                                & N                   & 97.62            & 86.02          & 87.68                             & \textbf{99.97}   & 98.6          & 91.54          & Own experiment                  \\
            \midrule
            \emph{BoW Models} \\
            SVM     & Y & --- & --- & --- & --- & 95\tnote{$6$} & --- & Silva et al.~\cite{DBLP:journals/air/SilvaCMW11} \\

            WideMLP                                 & N                   & 96.98            & 76.48          & 67.28                             & 99.86            & 97            & 82.26          & Own experiment                  \\
            fastText                                & N                   & 96.13            & 75.14          & 88.56\tnote{$4$}          & ---              & ---           & ---            & Zhao et al.~\cite{zhao2021sequential}      \\
            \midrule
            \textit{Graph-based Models} \\
            HGAT\tnote{$2$}                         & Y                   & ---              & 62.75          & 82.36                             & 63.21            & ---           & ---            & Yang et al.~\cite{hgat}                    \\
            NC-HGAT\tnote{$2$}                      & Y                   & ---              & 62.46          & ---                               & 63.76            & ---           & ---            & Sun et al.~\cite{nc-hgat}                 \\
            SGNN\tnote{$3$}                         & Y                   & 98.09            & 80.58          & 90.68\tnote{$4$}                  & ---              & ---           & ---            & Zhao et al.~\cite{zhao2021sequential}      \\
            ESGNN\tnote{$3$}                        & Y                   & 98.23            & 80.93          & \textbf{90.80}\tnote{$4$}         & ---              & ---           & ---            & Zhao et al.~\cite{zhao2021sequential}      \\
            C-BERT (ESGNN+BERT)\tnote{$3$}        & Y                   & 98.28            & 86.06          & 90.43\tnote{$4$}                  & ---              & ---           & ---            & Zhao et al.~\cite{zhao2021sequential}      \\
            DADGNN                                  & Y                   & 98.15            & 78.64          & ---                               & ---              & 97.99         & 84.32          & Liu et al.~\cite{liu2021deep}             \\
            DADGNN                                  & Y                   & 97.28            & 74.54          & 84.91                             & 98.16            & 97.54         & 82.81          & Own experiment                  \\
            HyperGAT                                & N                   & 97.97            & 78.32          & ---                               & ---              & ---           & ---            & Ding et al.~\cite{ding-etal-2020-hypergat} \\
            InducT-GCN                              & N                   & 96.68~\textsubscript{(0.13)}            & 75.34~\textsubscript{(0.35)}          & 76.67~\textsubscript{(0.45)}                             & 88.56~\textsubscript{(0.57)}            & 92.50~\textsubscript{(0.34)}          & 79.97~\textsubscript{(0.37)}          & Own experiment                  \\
            ConTextING-BERT                         & N                   & 97.91            & 86.01          & ---                                & ---              & ---           & ---           & Huang et al.~\cite{huang-etal-2022-contexting}                 \\
            ConTextING-RoBERTa                      & N                   & 98.13            & 89.43          & ---                                & ---              & ---           & ---           & Huang et al.~\cite{huang-etal-2022-contexting}                 \\
            \midrule
            \textit{CNN and LSTMs} \\
            SECNN\tnote{$3$}                        & Y                   & ---              & 83.89          & ---                               & ---              & 91.34         & 87.37          & Wang et al.~\cite{wang2021short}           \\
    MGNC-CNN & Y & --- & --- & --- & --- & 95.52 & 88.30\tnote{$7$} & Zhang et al.~\cite{DBLP:conf/naacl/ZhangRW16} \\ 
    DCNN     & Y & --- & 86.80\tnote{$8$} & --- & --- & 93 & --- & Kalchbr. et al.
    ~\cite{DBLP:conf/acl/KalchbrennerGB14} \\
            LSTM (BERT)                             & Y                   & 94.28            & 75.10          & 65.13                             & 99.83            & 97            & 81.38          & Own experiment                  \\
            Bi-LSTM (BERT)                          & Y                   & 95.52            & 75.30          & 66.79                             & 99.76            & 97.2          & 80.83          & Own experiment                  \\
            LSTM (GloVe)                            & Y                   & 96.34            & 74.99          & 67.67                             & 95.23            & 97.4          & 79.95          & Own experiment                  \\
            Bi-LSTM (GloVe)                         & Y                   & 96.84            & 75.32          & 68.15                             & 95.53            & 97.2          & 80.17          & Own experiment                  \\
            Bi-LSTM (GloVe)                         & Y                   & 96.31            & 77.68          & 84.81\tnote{$4$}                  & ---              & ---           & ---            & Zhao et al.~\cite{zhao2021sequential}      \\
            \toprule
            \textbf{Transductive Models}            & \textbf{Short?} & \textbf{R8}      & \textbf{MR}    & \textbf{Snippets}           & \textbf{Twitter} & \textbf{TREC} & \textbf{SST-2} & \textbf{Provenance}\\
            \midrule
            \textit{Graph-based Models} \\
            SHINE\tnote{$5$}                        & Y                   & ---              & 64.58          & 82.39                             & 72.54            & ---           & ---            & Wang et al.~\cite{shine}                   \\
            SHINE   & Y                   & 79.80~\textsubscript{(4.64)}            & 62.05~\textsubscript{(3.33)}         & 82.14~\textsubscript{(0.84)} & 70.64~\textsubscript{(1.95)}            & 79.90~\textsubscript{(4.06)}     & 61.71~\textsubscript{(1.34)}    & Own experiment                  \\
            STGCN                                   & Y                   & 97.2             & 78.2           & ---                               & ---              & ---           & ---            & Ye et al.~\cite{ye2020document}          \\
            STGCN+BiLSTM                            & Y                   & ---              & 78.5           & ---                               & ---              & ---           & ---            & Ye et al.~\cite{ye2020document}          \\
            STGCN+BERT+BiLSTM                       & Y                   & 98.5             & 82.5           & ---                               & ---              & ---           & ---            & Ye et al.~\cite{ye2020document}          \\
            SimpleSTC\tnote{$9$} & Y & --- & 62.27~\textsubscript{(1.11)} & 80.96~\textsubscript{(1.69)} & 62.19~\textsubscript{(1.56)} & --- & --- & Zheng et al.~\cite{DBLP:conf/emnlp/ZhengWYD22} \\
            TextGCN                                 & N                   & 97.07            & 76.74          & 83.49                             & ---              & ---           & ---            & Zhao et al.~\cite{zhao2021sequential}      \\
            TextGCN                                 & N                   & 97.07            & 76.74          & ---                               & ---              & 91.40         & 81.02          & Liu et al.~\cite{liu2021deep}             \\
            BertGCN                                 & N                   & 98.1             & 86.0           & ---                               & ---              & ---           & ---            & Lin et al.~\cite{lin2021bertgcn}          \\
            RoBERTaGCN                              & N                   & 98.2             & 89.7           & ---                               & ---              & ---           & ---            & Lin et al.~\cite{lin2021bertgcn}          \\
            TextGCN-BERT-serial-SB                  & N                   & 97.78            & 86.69          & ---                               & ---              & ---           & ---            & Zeng et al.~\cite{zeng2022simplified}      \\
            TextGCN-CNN-serial-SB                   & N                   & \textbf{98.53}   & 87.59          & ---                               & ---              & ---           & ---            & Zeng et al.~\cite{zeng2022simplified}      \\
            \bottomrule
        \end{tabular}
        } 
        
        \begin{tablenotes} 
            \item[$1$] With a batch size of 32 and for DeBERTa of 16.
            \item[$2$] With only 40 randomly selected documents per class.
            \item[$3$]Not reproducible. Authors have been contacted twice without a response.
            \item[$4$] Using a modified training split of $8,636$ training and $3,704$ test documents and further preprocessing.
            \item[$5$] Employing very low train ratios ($0.38\%$ to $6.22\%$).
            \item[$6$] Uni-gram model with extensive pre-processing, use of WordNet, etc. and 60 hand-coded rules\extended{ for the classification}
            \item[$7$] Removed phrases of length less than 4 from the training set
            \item[$8$] Using a slightly different split of 6,920 sentences for training, 872 for development, and 1,821 for test
            \item[$9$] Samples 20 labeled documents per class as training set and validation set
        \end{tablenotes}
    \end{threeparttable}
\end{table*}

\begin{table*}[!ht]
   \small
    \caption[Accuracy on our own short text classification datasets]{
    \centering
    Accuracy on our own short text classification datasets. The ``Short?'' column indicates whether the model makes claims about its ability to categorize short texts. Provenance refers to the source of the accuracy scores.
    }
    \label{tab:accuracy_own}
    \centering
    \begin{tabular}{lcllll}

        \toprule
        \textbf{Inductive Models} & \textbf{Short Text} & \textbf{NICE-45} & \textbf{NICE-2} & \textbf{STOPS-41} & \textbf{STOPS-2} \\
        \midrule
        \textit{Transformer Models} \\
        BERT                      & N                   & \textbf{72.79}   & 99.72           & 89.4              & 99.87            \\
        RoBERTa                   & N                   & 66.09            & \textbf{99.76}  & 89.56             & 99.86            \\
        DeBERTa                   & N                   & 59.42            & 99.72           & \textbf{89.73}    & 99.85            \\
        ERNIE 2.0                 & N                   & 45.55            & 99.69           & 89.39             & 99.85            \\
        ERNIE 2.0 (optimized)     & N                   & 67.65            & 99.72           & 89.65             & \textbf{99.88}   \\
        DistilBERT                & N                   & 69.28            & 99.75           & 89.32             & 99.85            \\
        ALBERTv2                  & N                   & 59.24            & 99.51           & 88.58             & 99.83            \\
        \midrule
        \textit{BoW Models} \\
        WideMLP                   & N                   & 58.99            & 96.76           & 88.2              & 97.05            \\
        \midrule
        \textit{Graph-based Models} \\
        DADGNN                    & Y                   & 28.51            & 91.15           & 26.75             & 97.48            \\ 
        InducT-GCN                & N                   & 47.06~\textsubscript{(0.61)}            & 94.98~\textsubscript{(0.23)}   & 86.08~\textsubscript{(0.07)}      & 97.74~\textsubscript{(0.06)}            \\
        \midrule
        \textit{CNN and LSTMs} \\
        LSTM (BERT)               & Y                   & 47.81            & 96.63           & 86.27             & 96.05            \\
        Bi-LSTM (BERT)            & Y                   & 52.39            & 96.63           & 85.93             & 98.54            \\
        LSTM (GloVe)              & Y                   & 52.64            & 96.17           & 87.4              & 99.46            \\
        Bi-LSTM (GloVe)           & Y                   & 55.35            & 95.93           & 87.38             & 99.43            \\
        \bottomrule
    \end{tabular}
\end{table*}

The accuracy scores for the text classification models on the six benchmark datasets are shown in Table~\ref{tab:accuracy}. The findings demonstrate that the relatively straightforward models LSTM, Bi-LSTM, and WideMLP provide a strong baseline across all datasets. 
This comparison clearly demonstrates the limitations of some models, with InducT-GCN falling short in all datasets except SearchSnippets, SECNN underperforming on TREC, and DADGNN producing weak MR results in our own experiment. 
The Transformer models, on the other hand, are the best performing across all datasets with the exception of SearchSnippets. With consistently strong performance across all datasets, DeBERTa stands out in particular.
The graph-based models from Zhao et el.~\cite{zhao2021sequential}, SGNN, ESGNN, and C-BERT, all perform well for the datasets for which results are available and ESGNN even outperforms all other models for SearchSnippets. 
It is important to note that Zhao et al.~\cite{zhao2021sequential} used a modified training split and additional preprocessing. 
While an increase of about $5$ percentage points for MR could be obtained by extending ESGNN with BERT in C-BERT, the increase is not noticeable for other datasets.
When applied to short texts, the inductive models even outperform transductive models. 
On Twitter, ERNIE 2.0 and ALBERTv2 reach a performance of $99.97\%$, and when using BERT on the TREC dataset, a performance of $99.4\%$ is obtained.
Non-Transformer models also perform well on TREC, although Transformers outperform them. 
For the graph-based models SHINE and InducT-GCN, we also calculated the mean and standard deviation of the accuracy scores across 5 runs. 
This is motivated from the observation that models based on graph-neural networks are susceptible to the initialization of the embeddings~\cite{pitfallsShchur2018}.
SHINE had a generally high standard deviation of up to nearly 5 points, indicating greater variance in its performance.
In comparison, InducT-GCN has a rather small variance of always below 1 point. 

The accuracy results for our newly introduced datasets, NICE and STOPS, are shown in Table~\ref{tab:accuracy_own}.
New characteristics covered by NICE and STOPS include shorter average lengths and the ability to distinguish between classes at a fine-granular level in NICE-45 and STOPS-41.
The investigation of more documents is also conducted in the case of STOPS.
As a result, NICE-45 and STOPS-41 reveal that DADGNN encounters issues when dealing with more classes, even  falling around $20$ and $60$ percent points behind the baseline models.
While still performing worse than the baseline models, InducT-GCN outperforms DADGNN on all four datasets.
Transformers once again demonstrate their strength and rank as the top performing models across all datasets on this dataset. There are also significant drops. ERNIE 2.0 performs worse than the baseline models with $45.55\%$ on NICE-45. 
However, ERNIE 2.0 (optimized), which uses different hyperparameter values (see Section~\ref{sec:hyperparameter}), comes in third with $67.65\%$

\section{Discussion}
\label{sec:discussion}

Graph-based models are computationally expensive because they require not only the creation of the graph but also its training, which can be resource- and time-intensive, especially for word-document graphs with $\mathcal{O}(N^2)$ space \cite{galkeMLP}.
On STOPS, this drawback becomes very apparent. We could observe that DADGNN required roughly $30$ hours of training time, while BERT only took $30$ minutes to fine-tune with the same resources.
Although in the case of BERT, the pre-training was already very expensive, transfer learning allows this effort to be used for a variety of tasks.
Nevertheless, the Transformers outperform the inductive graph-based models as well as the short text models, with just one exception.
The best model for SearchSnippets is ESGNN, but additional preprocessing and a modified training split were employed.
Our Bi-LSTM results, obtained without additional preprocessing, differ by $16.66$ percentage points from the Bi-LSTM results from Zhao et el.~\cite{zhao2021sequential}. 
This indicates that preprocessing, and not a better model, is primarily responsible for the strong outcomes of the SearchSnippets experiments.
Another interesting discovery can be made using the sentiment datasets. In comparison to other datasets, the Transformers outperform graph-based models that do not utilize a Transformer themselves by a large margin. This demonstrates that graph-based models may not be as effective at sentiment prediction tasks.
In contrast, the CNN-based models show strong performance on the sentiment analysis task SST-2.
Still, the best CNN model is more than 6 points below the best transformer.
However, it should be noted that not all Transformers are consistently excellent.
For instance, for NICE-45, one can observe a lower performance with ERNIE 2.0. 
But the absence of this performance decrease in our optimized version of ERNIE~2.0~(optimized) suggests that choosing suitable hyperparameters is crucial in this case.

\subsection{Key Results}
\label{sec:keyresults}

Our experiments unambiguously demonstrate that Transformers achieve \ac{SOTA} accuracy on short text classification tasks. This raises the question of whether specialized short text techniques are necessary given that the performance of the existing models is insufficient.
This observation is especially interesting because many of the short text models used are from $2021$~\cite{hgat,zhao2021sequential,liu2021deep,wang2021short} or $2022$~\cite{nc-hgat}.
Most short text models attempt to enrich the documents with some kind of external context, such as a knowledge base or \ac{POS} tags. However, one could argue that Transformers implicitly contain context in their weights through their pre-training.

Those short text models that compare themselves to Transformers assert that they outperform them.
For instance, Ye et al.~\cite{ye2020document} claim to outperform BERT by $2.2$ percentage points on MR, but their fine-tuned BERT only achieves $80.3\%$.
In contrast, our own experiments show that BERT achieves $86.94\%$.
With $85.86\%$ on MR, Zhao et al.~\cite{zhao2021sequential} achieve better BERT results, but only to beat it by a meager $0.2\%$ with C-BERT. Given the low surplus, they would no longer outperform it with a marginally better selection of hyperparameters for BERT.
Therefore, it is reasonable to assume that the importance of good hyperparameters for Transformers is underestimated and that they are often not properly optimized.
ERNIE~2.0~(optimized), which outperforms ERNIE~2.0 on every dataset, also demonstrates the effect of better hyperparameters.
Finally, Zhao et al.~\cite{zhao2021sequential} is already outperformed by other transformers like RoBERTa and DeBERTa by 3 and 4 points, respectively.

Additionally, there is a need for new short text datasets because the widely used benchmark datasets share many characteristics and fall short in many use cases.
The common benchmark datasets all contain around $10,000$ documents, distinguish only a few classes, and frequently have a similar average length. Furthermore, many of them cover the same tasks. For instance, MR, Twitter, and SST-2 all perform sentiment prediction, which makes sense given how much short text is produced by social media.
In this paper, we introduce two new datasets with distinctive attributes to cover more cases in NICE and STOPS. 
New and intriguing findings are produced by the new characteristics that are investigated using these datasets.
Particularly, the ability to distinguish between classes at a fine-granular level reveals the shortcomings of various models, like DADGNN or ERNIE 2.0.
NICE-45 in particular proved to be challenging for all models, making it a good benchmark for future advancements.

\subsection{Threats to Validity}
\label{sec:threattovalidity}

In our study, each experiment was generally conducted once.
The rationale is the extremely low standard deviation for text classification tasks observed in previous studies~\cite{galkeMLP,zhao2021sequential,liu2021deep}.
\extended{It raises the prospect that multiple runs are
 unnecessary and does not justify the higher computational expenses.}  
However, it has been reported in the literature on models using graph neural networks (GNN) that they generally have high standard deviation in their performance, which has been attributed among others to the influence of the random initialization in the evaluation~\cite{pitfallsShchur2018}. 
Thus, we have run our experiments for SHINE and InducT-GCN five times and report averages and standard deviation. 
The high standard deviation observed in SHINE's performance adds to the evidence of the need for caution when interpreting the results of \acp{GNN}~\cite{pitfallsShchur2018}\extended{ and underscores the importance of conducting multiple runs and reporting the mean and standard deviation}.

 We acknowledge that STOPS contains user-generated labels, some of which may not be entirely accurate. However, given that this occurs frequently in numerous use cases, it is also crucial to test the models in these scenarios.

\subsection{Parameter Count of Models}
Table~\ref{tab:num_params} lists the parameter counts of selected Transformer models, the BoW-based baseline methods WideMLP, and graph-based methods used in our experiments.
Generally, the top performing Transformer models have a similar size between 110M to 130M parameters. 
Although DistilBERT is only have of that size and ALBERTv2 only about a tens, our experiments show still comparable accuracy scores on R8, Snippets, Twitter, and TREC.
ALBERTv2 with its 12M parameters outperforms the WideMLP baseline with 31.3M parameters on all datasets, some with a large margin.
The graph-based model ConTextING-RoBERTa has a similar parameter count compared to the pure Transformer models, since the RoBERTa transformer is used internally.
It is the top-performer among the graph-based models on R8 and MR but cannot outperform the pure Transformer models.

\begin{table}[ht]
   \small
    \centering
    \caption{Parameter counts for selected methods used in our experiments}\label{tab:num_params}
    \begin{tabular}{lr}
    \toprule
    \textbf{Model} & \textbf{\#parameters}  \\
    \midrule

\textit{Transformer models} &  \\
         BERT & 110M\\
         RoBERTA  & 123M\\
         DeBERTA  & 134M\\
         ERNIE 2.0 & 110M \\   
         DistilBERT  & 66M\\
         ALBERTv2 & 12M \\         

    \midrule

    \textit{BoW-based methods} & \\
         WideMLP  & 31.3M\\
    \midrule

    \multicolumn{2}{l}{    \textit{Graph-based methods}  }\\
        HyperGAT & LDA parameters + 3.1M\\
         ConTextING-RoBERTa & 129M \\

     \bottomrule
    \end{tabular}
\end{table}

\subsection{Generalization}
\label{sec:generalization}

As we cover in our experiments a range of diverse domains, with sentiment analysis on various themes (MR, SST-2, Twitter), question type classification (TREC), news (R8), and even search queries (SearchSnippets), we expect to find equivalent results on other short text classification datasets.
Additionally, the categorization of goods and services is covered by our new datasets NICE and STOPS. They include additional features not covered by the benchmark datasets, including a significantly larger amount of training data in STOPS, a shorter average length, and the capacity to differentiate between a wider range of classes.
By using an example from a business problem, STOPS specifically demonstrates how the knowledge gained here can be applied in corporate use.

In this work, we cover a variety of models for each architecture, particularly the most popular and best-performing ones.
Our findings are consistent with the studies by Galke and Scherp~\cite{galkeMLP}, which demonstrate the tremendous power of Transformers for traditional text classification.

\section{Conclusion and Future Work}
Our experiments unequivocally demonstrate the outstanding capability of Transformers for short text classification tasks.
Additional research on our newly released datasets, NICE and STOPS, supports these findings and highlights the issue of becoming overly dependent on benchmark datasets with a limited number of characteristics.
In conclusion, our study raises the question of whether specialized short text techniques are required given the lower performance of current models.

Future research on improving the performance of Transformers on short text could be to do pre-training on short texts or on in-domain texts (\ie pre-training in the same domain as the target task) \cite{sun2019fine,oreillyTransformer,brinkmann2021improving}, multi-task fine-tuning \cite{sun2019fine,oreillyTransformer}, or an ensemble of multiple Transformer models \cite{BERTensembleRanking}. 

\paragraph*{Acknowledgement.}
This work is co-funded under the 2LIKE project by the German Federal Ministry of Education and Research (BMBF) and the Ministry of Science, Research and the Arts Baden-Württemberg within the funding line Artificial Intelligence in Higher Education.
We thank Till Blume and Felix Krieger from Ernst \& Young (EY) for the discussion of the problem statement that motivated this work.
We are grateful to Liu et al.~\cite{liu2021deep} for providing the unreleased DADGNN source code.

\newpage
\bibliographystyle{splncs04}
\bibliography{library}

\end{document}